\documentclass[a4paper,10pt, twocolumn]{article}
\pdfoutput=1
\newcommand{\citep}[1]{\cite{#1}}

\usepackage{times}
\usepackage{amsmath}
\usepackage{amsfonts}
\usepackage{amssymb}
\usepackage{savesym}
\usepackage{graphicx}
\usepackage{epstopdf}
\savesymbol{AND}
\savesymbol{OR}
\savesymbol{NOT}
\savesymbol{TO}
\savesymbol{COMMENT}
\savesymbol{BODY}
\savesymbol{IF}
\savesymbol{ELSE}
\savesymbol{ELSIF}
\savesymbol{FOR}
\savesymbol{WHILE}
\usepackage{algorithmic}
\usepackage{color}
\usepackage[normalem]{ulem}

\usepackage[inline]{enumitem}
\usepackage[font={small,it}]{caption}
\usepackage{algorithm}
\usepackage{booktabs}
\usepackage{siunitx}
\usepackage[numbers]{natbib}
\usepackage[top=1in, bottom=1in, left=0.85in, right=0.85in]{geometry}

\usepackage{amsmath}
\newcommand{\differential}{\text{d}}

\DeclareMathOperator*{\argmin}{arg\,min}

\title{\LARGE \bf RRT-CoLearn: towards kinodynamic planning without numerical trajectory optimization}

\author{W.J.Wolfslag\thanks{Department of 3ME, Delft University of Technology} \thanks{Corresponding author: 2628CD-2 Delft NL, w.j.wolfslag@tudelft.nl}, M.Bharatheesha\footnotemark[1], T.M.Moerland\thanks{Department of Computer Science, Delft University of Technology}, M.Wisse\footnotemark[1] \thanks{This  work  is  part  of  the  research  program  STW,  which  is
(partly)  funded  by  the  Netherlands  Organization  for  Scientific Research (NWO). This work  is also part of the European Community’s Seventh  Framework  Programme  (FP7/2007-2013)  under  grant agreement No. 609206.}}
\date{}

\begin{document}
\maketitle

\begin{abstract} \bf
Sampling-based kinodynamic planners, such as Rapidly-exploring Random Trees (RRTs), pose two fundamental challenges: computing a reliable (pseudo-)metric for the distance between two randomly sampled nodes, and computing a steering input to connect the nodes. The core of these challenges is a Two Point Boundary Value Problem, which is known to be NP-hard. Recently, the distance metric has been approximated using supervised learning, reducing computation time drastically. The previous work on such learning RRTs use direct optimal control to generate the data for supervised learning. This paper proposes to use indirect optimal control instead, because it provides two benefits: it reduces the computational effort to generate the data, and it provides a low dimensional parametrization of the action space. The latter allows us to learn both the distance metric and the steering input to connect two nodes. This eliminates the need for a local planner in learning RRTs. Experimental results on a pendulum swing up show 10-fold speed-up in  both the offline data generation and the online planning time, leading to at least a 10-fold speed-up in the overall planning time.
\end{abstract}

\section{Introduction}
\label{sec:introduction}

For motion planning of robotic manipulators, kinodynamic planning and sampling-based planning are getting increasingly popular. Kinodynamic planning, i.e., planning in state-space rather than configuration space, improves robustness, speed and energy efficiency of robots~\cite{Collins2005, Wolfslag2015, Plooij2016}. Sampling based planning has been shown to be the most viable way to handle high dimensional spaces and obstacles~\cite{Hsu2002,LaValle2001}. In this paper, we will consider how to apply Rapidly-exploring Random Trees~\cite{LaValle2001}, the most popular sampling-based planning algorithm, to kinodynamic planning.

\begin{figure}[t]
\includegraphics[width=1.0\columnwidth]{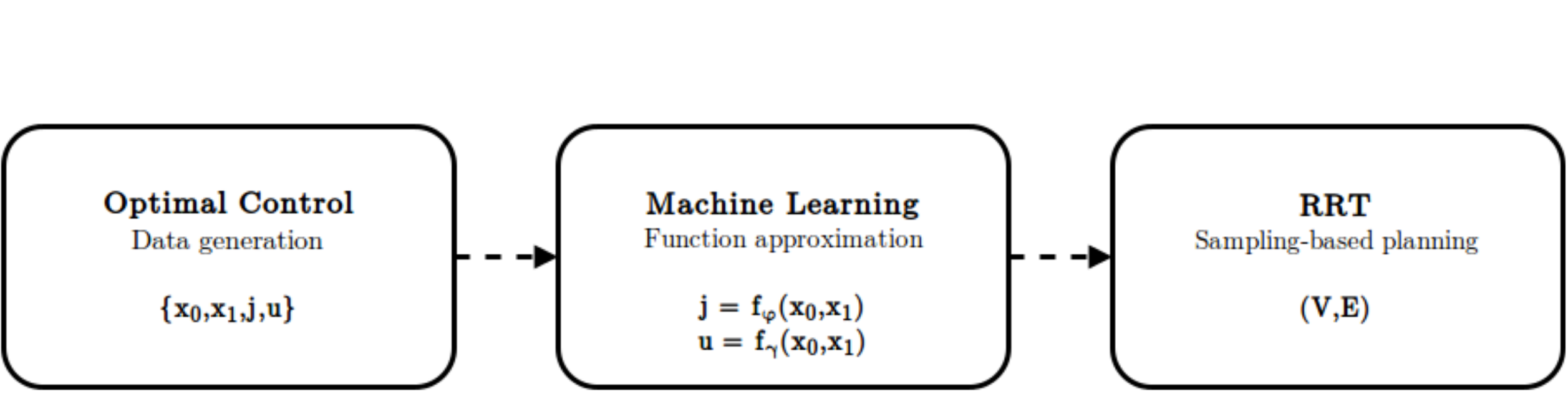}
\caption{Schematic illustration of the Learning-RRT architecture. First, we use optimal control to generate a dataset which, for a given start state $x_0$ and end state $x_1$, specifies the cost-to-go $j$ and control input $u$. Subsequently, we use machine learning to predict the cost-to-go function parametrized and required control input given a start and goal node. These first two (computationally intensive) phases happen {\it offline}. Subsequently, we input the function approximators to the RRT for {\it online} planning. The function approximators provide the RRT with quick cost prediction and avoids the need for optimization in a local planner.} \label{schema}
\end{figure}

RRT builds a tree graph structure with the states of the system as tree nodes and the trajectories of the system between two states as tree edges. The algorithm selects a node to expand from based on a \emph{distance} to a randomly sampled node in the state space, i.e., it selects the \emph{nearest} node in the current tree. The algorithm then expands that tree node, by means of a local planner that aims to reach the randomly-sampled node. These steps happen online, so computation time is crucial. Unfortunately, in state-space, both local planning and distance computation are computationally expensive~\cite{LaValle2001}. One approach to reduce the computational burden is to approximate the true distance function by a heuristic~\citep{Glassman2010,Perez2012, Karaman2013, Shkolink2009}. Two frequently used heuristics are the Euclidean distance and the optimal cost-to go of a linear approximation to the system. The convergence of RRT variants using the Euclidean distance heuristic, and random steering inputs, was extensively analysed in~\cite{Li2016}. For promising results for the linearizing heuristic see~\citep{Goretkin2013, Webb2013, Schmerling2015}. However, these heuristics only minimally utilize the dynamical properties of the system, and therefore typically require more nodes to solve a given problem using RRT.

Another approach, which we call Learning-RRT, was proposed recently~\cite{Bharatheesha2014,Palmieri2014} and has already shown promising initial results~\cite{Allen2016}. Learning-RRT involves an offline machine learning phase that learns the distance and steering function in the RRT~(Figure~\ref{schema}). An optimal control algorithm provides a database of optimal trajectories, which is the input for a supervised learning algorithm. This algorithm learns to approximate the functions the RRT requires. Supervised learning provides two benefits: 1) generalization over state-space and 2) fast online predictions. Thereby, the computational burden of trajectory optimization does not have to be repeated for new situations, and, it is shifted offline.

Note that the final trajectory found by the RRT-algorithm is in general not optimal, even though the trajectories in the database are optimal. The database therefore is not build up of optimal trajectories to improve the cost to go of the RRT-trajectory. Rather, optimality improves learning performance by providing a meaningful cost-to go metric and trajectories that tend to have a similar shape if they connect similar points in state space.

This paper proposes a method that helps to overcome the two remaining challenges of Learning-RRT:
\begin{enumerate}
\item{{\bf Local planning}: In literature on learning-RRT, only the distance function is approximated by machine learning. Recall that the RRT also requires a steering function. Supervised learning of that function is hard due to the large number of parameters typically required to describe optimal input signals. Therefore, previous Learning-RRTs resort to a computationally expensive numerical optimization for their steering function~\cite{Bharatheesha2014}.}
\item{{\bf Dataset generation}: The dataset for the supervised learning algorithm consists of many optimal trajectories. As optimizing a single trajectory already is a significant computational burden, generating a full dataset is very computationally demanding.}
\end{enumerate}

The main contribution of this paper is the use of indirect optimal control to generate the dataset. This allows us to solve both the problems mentioned above, thereby decreasing the computational burden by up to two orders of magnitude, both in the offline and the online phase of the learning RRT planning. However, the dataset generated by this method contains a bias, which is problematic for the learning algorithm. It turns out we can efficiently remove this bias through a simple dataset resampling algorithm.

The focus of this paper is to introduce this novel method and its implementation details. Our experiments provide proof of concept by evaluating our method on the pendulum swing-up task studied in the previous learning-RRT implementation by Bharetheesha et al.~\cite{Bharatheesha2014}. This allows us to demonstrate and compare the validity of our method. To our knowledge, we are the first to demonstrate learning of the control input in a kinodynamic sampling-based planning task.

The structure of this paper is as follows. We start with a generic specification of the Learning-RRT algorithm (Section \ref{sec:learningRRT}). This specification is applicable to any data generation method, and intended to structure all Learning-RRT components. Subsequently, the main contribution follows in Section \ref{sec:dataGeneration}, which tackles the problems of \textit{local planning} and \textit{dataset generation}. Then we discuss how to remove the dataset bias introduced by optimal control in Section \ref{sec:datasetCleaning}. In Section \ref{sec:errors}, we discuss how the cost-to go metric, and the learned approximation thereof, affect the convergence of the RRT algorithm. The experimental proof of concept of our algorithm is presented in Section \ref{sec:Experiments}. Finally, Sections \ref{sec:discussion} and \ref{sec:conclusion} contain discussion and conclusions.

\section{Learning-based RRT}
\label{sec:learningRRT}
Learning-based RRTs leverage the benefits of (supervised) learning to speed up the computationally expensive modules of a kinodynamic RRT. The Learning-RRT algorithm is presented in Algorithm \ref{alg:rrt_learning}.

\begin{algorithm}[t]
\small
\caption{Learning RRT ((V, E), N)}
\begin{algorithmic}
\label{alg:rrt_learning}
\STATE $\hat{D} \leftarrow \texttt{generate\_data}(N) \quad$// Section \ref{sec:dataGeneration}
\STATE $D \leftarrow \texttt{clean\_data}(\hat{D}) \quad$ // Section \ref{sec:datasetCleaning}
\STATE $\hat{J}\leftarrow \texttt{fit\_cost}(D)$
\STATE $\hat{U} \leftarrow \texttt{fit\_input}(D)$
\STATE $\hat{V}\leftarrow\texttt{fit\_valid}(D) \quad$ // Section \ref{sec:errors}
\STATE $(X,E) \leftarrow(x_{\text{initial}},\emptyset)$
\STATE solutionfound $\leftarrow$ False
\WHILE{NOT(solutionfound)}
\STATE $x_\text{target}\leftarrow \texttt{sample}()$
\IF{$ANY(\hat{V}(x,x_\text{target})) \forall x \in X$}
\STATE $x_\text{nearest}\leftarrow\argmin_{x\in X}\hat{J}(x,x_\text{target})$
\STATE $\left(c,x,u\right)\leftarrow \texttt{simulate}(x_\text{nearest},\hat{U}(x_\text{nearest},x_\text{target}))$
\STATE $X\leftarrow X\cup \{x\}$
\STATE $E\leftarrow E\cup \{\left(x_\text{nearest},\,x,\,c\right)\}$
\ENDIF
\ENDWHILE
\RETURN $X,E$
\end{algorithmic}
\end{algorithm}

The first step in the algorithm is to create a dataset of optimal trajectories. Specifically, we generate a dataset $D = \{b^i\}_{i=1}^N$, where each entry $b^i = \{x^i_0,x^i_1,j^i,u^i\}$ consists of an initial state $x_0\in \mathcal{X}$, a final state $x_1\in \mathcal{X}$, a distance metric/cost-to-go $j\in \mathbb{R}^{+}$, and a set of parameters $u \in \mathcal{U}$, that describe the optimal input leading the system from state $x_0$ to state $x_1$. Note that $\mathcal{U}$ can take many forms, depending on the discretization used. For example, in previous work  the input has been cast as a polynomial over time, with $\mathcal{U}$ being the coefficients of that polynomial~\citep{patterson2014gpops}. Alternatively, when the input is cast as a piecewise-linear function, $\mathcal{U}$ consists of the values of the function at the switch-times.

The optimal trajectories are generally found using a numerical algorithm searching for local optima, which poses a challenge for the supervised learning algorithm. If two solutions are nearby in $x_0$ and $x_1$, but hail from a different local optimum, a deterministic supervised learning algorithm (for example trained on mean-squared error) will predict the average over the two solutions. This not only makes the cost prediction inaccurate, but most importantly ruins the steering input prediction: the average of the two steering inputs in the data could lead to a completely different state than the target state. This local-optimum bias requires us to create the dataset $D$ in two stages. The first stage creates a dataset $\hat{D}$ of size $N$ which contains local-optimum-bias, and is indicated in Alg.~\ref{alg:rrt_learning} by the function \texttt{generate\_data}. The second stage \texttt{clean\_data} removes the local-optimum-bias to create the desired dataset $D$.

The second step is to use a supervised learning algorithm on $D$ to approximate the two functions that define the optimal control solution: \begin{enumerate*}\item the function $\hat{J}: (\mathcal{X},\,\mathcal{X})\rightarrow \mathbb{R}^{+}$, which maps from an initial and a final state to a cost-to-go, \item the function $\hat{U}: (\mathcal{X},\,\mathcal{X})\rightarrow\mathcal{U}$, mapping the initial and final state to the required input parameters.\end{enumerate*}

For both function approximators we implement k-nearest neighbours \citep{friedman2001elements}, a standard non-parametric function approximator with robust performance in smaller state-spaces. For a dataset test point $x$, we identify the $k$ nearest neighbours in our dataset $D$ based on Euclidean distance. The predicted value (e.g. for cost $j$) for the test point then becomes the average value of these neighbours. We cover possible extensions to other supervised learning techniques in the Discussion.

The next step  in the algorithm, \texttt{fit\_valid}, addresses some inherent limitations of supervised learning approaches. We defer further details on this step to Section \ref{sec:errors}.

The fourth stage of the Learning-RRT algorithm is essentially the online RRT stage. First, sample a point in statespace. Then test if there is a valid connection from any node in the tree to the sampled point. If that is the case, expand the node that is nearest to the sampled point according to the function $\hat{J}$. The expansion will use the learned input parameter function $\hat{U}$, which will likely make a small error. The new node is therefore not exactly the sampled point, but a point in statespace that hails from the approximation $\hat{U}$. The algorithm iterates these steps until it connects to the desired region in state-space. Finally, as standard in RRTs the sampling of new nodes includes a goal bias: it will intermittently replace the uniform state-space sample with the desired end-point.

\section{Data generation}
\label{sec:dataGeneration}

The dataset for the function approximator is generated from a set of optimal trajectories for the system and cost function under consideration. The most common approach to find these trajectories are the so-called \emph{direct} optimal control approaches~\cite{patterson2014gpops,posa2014direct,Tassa2012}.
In these approaches the state equations and cost function are approximated by a discretized system, which is then numerically optimized.

An alternative to direct optimal control is the much older \emph{indirect} approach \citep{rao2009survey,pontryagin1987mathematical}, which first optimizes and then discretizes. For many applications, direct approaches replaced the indirect approach due to better numerical stability at long planning distances. However, it turns out indirect optimal control is ideally suited for the RRT scenario. First, the numerical instability poses no problem for the short segments that are required for RRT. Furthermore, indirect optimal control brings two important benefits. First, it provides a parametrized control input in low dimensional space, which allows learning of the control input. Second, it removes the need for optimization in the sampling process, which speeds up data generation. We will explain both benefits at the end of this section, after introducing the indirect optimal control method. At that point, we will also explain the remaining downside of the indirect optimal control approach: a more biased dataset.

\subsection*{Indirect optimal control}
This section will discuss the standard indirect optimal control procedure, which forms the basis for our dataset generation method.
The exposition derives the optimal control equations for our experimental system: the single pendulum swing-up. The procedure for other systems will mostly follow the same outline; small differences can occur. For those differences, more details and proofs we refer to \citep{naidu2002}.

The optimal control approach aims to find the functions $x(t)$ and $u(t)$ from time $t\in\mathbb{R}$
 to state $x\in\mathbb{R}^n$ and input $u\in\mathbb{R}^m$, that minimizes a cost function of the following form:
\begin{equation}
J(x(t),u(t))=\int_0^{t_\text{f}}C(x(t),u(t))\differential t \label{eq:optimalCost}
\end{equation}
Subject to the constraints:
\begin{align}
&\dot{x}(t) = f(x(t),u(t))\quad \forall t\in(0,t_\text{f}),\nonumber\\
&x(0) = x_{\text{initial}}, \quad x(t_\text{f}) = x_{\text{final}}
\end{align}
where $x_\text{initial}$ and $x_\text{final}$ are fixed initial and goal states, and the final time $t_f$ is optimized along with the trajectory and input function. In the remainder we will often drop the explicit dependency on the time $t$.

For the single pendulum we have $x = (\theta,\omega)$, where $\theta$ and $\omega$ are the (angular) position and velocity respectively. With $u$ being a torque, we get $f = (\omega, \sin(\theta) + u)$. Finally, as a cost function, we take into account both the time it takes to reach the goal-state, as well as the energy expended to get there, by setting the cost integrand to $C = w + u^2/2$, with $w$ a weight which tunes the contribution of the time component in the cost function.

The first step in the indirect optimal control approach is to define the Hamiltonian $\mathcal{H}$, which is the sum of the integrand $C$ and the inner product of  a vector of Lagrange multipliers with the state equations. The Lagrange multipliers are called the costates, and in the case of the pendulum consist of $(\lambda_\theta,\lambda_\omega)$. We then get the Hamiltonian:
\begin{equation}
\mathcal{H}(x,\lambda,u) = w + \frac{u^2}{2} + \lambda_\theta\omega + \lambda_\omega(\sin(\theta) + u)
\end{equation}

The second step is finding an optimal input $u^*$, by minimizing the Hamiltonian with respect to the input:
\begin{equation}
u^* = \argmin_{u}\mathcal{H}=-\lambda_\omega
\end{equation}
The third step is creating the optimal Hamiltonian, by replacing the input with the optimal input:
\begin{equation}
\mathcal{H}^*(x,\lambda) = w + \lambda_\theta\omega + \lambda_\omega\sin(\theta) -\frac{\lambda_\omega^2}{2}
\end{equation}
The fourth step computes a system of ordinary differential equations (ODEs) that specify the evolution
of the optimal state and costate over time :
\begin{align}
&\quad\dot{\theta} = \frac{\partial{\mathcal{H}^*}}{\partial{\lambda_\theta}} = \omega
\quad\quad\quad\quad\dot{\omega} = \frac{\partial{\mathcal{H}^*}}{\partial{\lambda_\omega}} = \sin(\theta) -\lambda_\omega \label{eq:state}\\
&-\dot{\lambda}_\theta= \frac{\partial{\mathcal{H}^*}}{\partial{\theta}} = \lambda_\omega\cos(\theta) \quad\quad
-\dot{\lambda}_\omega = \frac{\partial{\mathcal{H}^*}}{\partial{\omega}} = \lambda_\theta \label{eq:costate}
\end{align}

In typical use of the indirect optimal control approach, the last step is to use the Equations~\ref{eq:state}-\ref{eq:costate} to find the optimal trajectory. For a given costate, the above system of equations are (numerically) integrated, which results in a locally optimal state trajectory. Note that this trajectory depends on the choice of initial costate, and the time duration of the integration. By tuning the initial costate and final time, we find a locally optimal state trajectory that reaches the desired state. This tuning normally requires a numerical optimization method, which minimizes the difference between final state and desired state. In the next section, we will show that such an optimization is not required for us.

Optimal control problems solved for Learning-RRTs typically have a free final time and a cost integrand $C$ that does not explicitly depend on time. For these problems there is an additional constraint on the initial costate.
\begin{equation}
\mathcal{H}^*(x(0),\lambda(0)) = 0 \label{eq:costateConstraint}
\end{equation}

The main reason why \emph{indirect} optimal control is largely replaced by direct methods, such as multiple shooting, is that the resulting differential equations are unstable, and therefore difficult to numerically solve reliably.  However, the RRT algorithm splits the whole motion into small parts, so it is only required to solve problems with a small integration period, making this instability less important.

\subsection*{Benefits of using indirect optimal control}
We can immediately see two major advantages of this optimal control approach for use in Learning-RRT setting. First, the input directly follows from the costates, i.e., in principle, there is a mapping $U(x_\text{initial},x_\text{final}) \rightarrow (\lambda_\theta(0),\lambda_\omega(0),t_\text{f})$, from initial and final state to a set of only three parameters that describe the input function. This set is small, and will only grow linearly with the size of the statespace (and is even independent of the number of inputs). In contrast, direct optimal control approaches require to parametrize inputs as functions over time, which results in much larger spaces of parameters to learn. The reduction in the number of parameters means the input function can be learned efficiently, thereby solving the problem of \emph{local planning} in RRTs as identified in the introduction.

To see the second major advantage, look at how the whole dataset is created. In current learning RRTs, the dataset is generated by sampling from the $(x_\text{initial},x_\text{final})$-space, and then, find an optimal trajectory and cost for each sample. Note that every combination of initial state, initial costate and final time produces an optimal trajectory for a certain final state. So, if we sample from the allowed initial states, initial costates and final times, we effectively sample over all initial state - final state combinations. While previous approaches had to numerically optimize the steering input, we can eliminate the need for numerical optimization by  sampling the costate, meaning the data are generated much faster.

Algorithm \ref{alg:dataGeneration} outlines the data generation procedure. The functions \emph{random\_State}, \emph{random\_Costate}, and \emph{random\_Time} specify random states, costates and final times within bounds that depend on the problem. Furthermore, the function \emph{random\_Costate} takes into account Eq.~\ref{eq:costateConstraint}. Finally, the function \emph{integrate} numerically integrates the optimal control equations until the final time. To optimize the efficiency of the data generation we incorporate the intermediate integration results in the data set as well, even though the resulting datapoints are therefore no longer independent from each-other.

In this section, we have outlined the indirect optimal control approach to solving the \emph{data generation} problem in the learning RRT algorithm. Because the optimal control algorithm incorporates {\it costates}, we name the overall algorithm RRT-CoLearn. We have also discussed its two major advantages: learning optimal steering inputs (increasing online speed) and generating data without optimizing (increasing offline speed).

\begin{algorithm}[t]
\small
\caption{$\text{generate\textunderscore data(N)}$}
\begin{algorithmic}
\label{alg:dataGeneration}
\STATE Optimal\_ODEs $\leftarrow$ Eqs.~\ref{eq:optimalCost}-\ref{eq:costate}
\STATE $\hat{D}$ $\leftarrow$ empty()
\FOR {$n = 1:N$}
\STATE $x_\text{initial}$ $\leftarrow$ random\_State()
\STATE $\lambda_\text{initial}$ $\leftarrow$ random\_Costate() s.t. Eq.~\ref{eq:costateConstraint}
\STATE $T_\text{final}$ $\leftarrow$ random\_Time()
\STATE $x_\text{final}$, $J$ $\leftarrow$ integrate(Optimal\_ODEs,$x_\text{initial}$ ,$\lambda_\text{initial}$,$T_\text{final}$)
\STATE append($\hat{D}$,\{$x_\text{initial}$,$x_\text{final}$,$J$,$\lambda_\text{initial}$\})
\ENDFOR
\RETURN $\hat{D}$
\end{algorithmic}
\end{algorithm}

\section{Dataset cleaning}
\label{sec:datasetCleaning}

\begin{figure}
\includegraphics[width=\columnwidth]{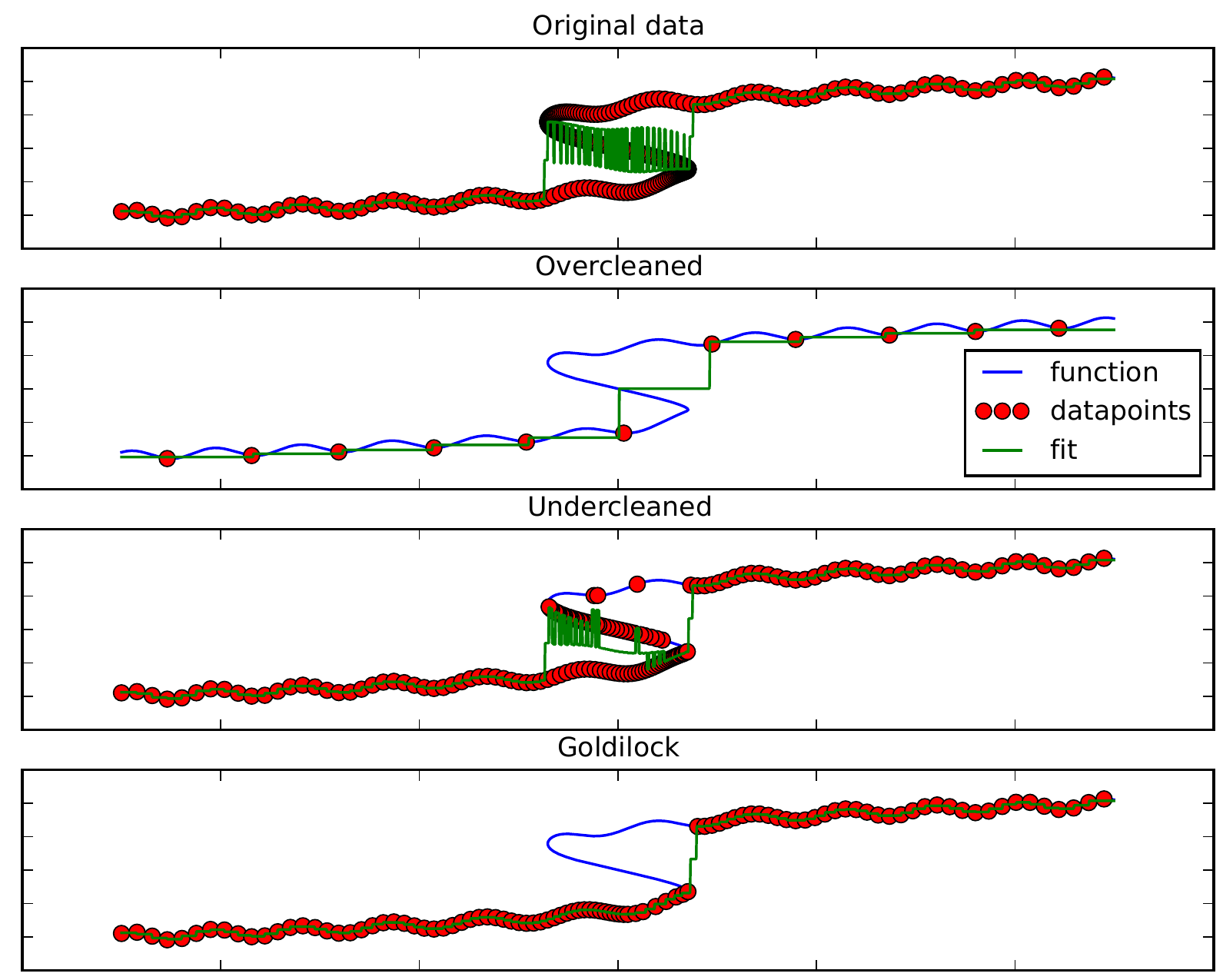}
\caption{The data-bias problem and the effect of the $d$ parameter in the data cleaning algorithm. The top figure shows an imaginary dataset, which has a problem with bias in the middle of its domain. The fitted function is a poor approximation of the least cost part of the datapoints. The second figure shows a cleaned dataset where the value for $d$ was chosen too large: the bias is gone, but there is not enough resolution left to accurately fit the function. In the third function $d$ is chosen to small: not all bias is removed. Finally, the bottom figure shows a cleaned dataset with a proper choice for $d$: the bias is removed, and enough resolution remains.}
\label{fig:distanceEffect}
\end{figure}

The dataset generated by Algorithm \ref{alg:dataGeneration} originates from a search for local optima, and can therefore include a bias that interferes with learning performance. The problem is illustrated for with an artificial dataset in Figure \ref{fig:distanceEffect} (top), where in the middle input region we have the global optimum at the bottom, but there are local optima above it. A standard function approximator (with squared loss) for a given point in input space (independent variable) predicts the expectation of the dependent variable. This results in the conditional mean of the datapoints, as shown by the green line in Figure \ref{fig:distanceEffect} (top). Note how the predicted function deteriorates in the middle segment, where it predicts the average instead of the bottom (optimal) cost.

This is problematic for predicting the cost function and especially harmful for predicting the control parameters. Averaging over two locally optimal control inputs by no means guarantees that we end up anywhere close to the target. We therefore need a dataset cleaning algorithm, i.e., a procedure that somehow eliminates the non-optimal datapoints. In literature, there are resampling methods for dataset imbalance, most noteworthy class label imbalance in classification tasks ~\citep{galar2012review}. However, our dataset is not imbalanced, but rather contains a systematic bias. It turns out we can leverage the fact that the noise is systematic to come up with a simple resampling/cleaning algorithm.

\begin{algorithm}[t]
\small
\caption{clean\_data($\hat{D},d,k_\text{max}$)}
\begin{algorithmic}
\label{alg:clean_data}
\STATE $D \leftarrow \hat{D}$
\STATE $k \leftarrow 0$
\WHILE {$k < k_\text{max}$}
\STATE $p_\text{sample} \leftarrow$ selectRandom($D$)
\STATE $p_\text{neigh} \leftarrow$ nearestNeighbour($p_\text{sample},D$)
\IF {distance($p_1,p_2)<d$}
\STATE $k \leftarrow 0$
\STATE $p_\text{high} \leftarrow \argmin_{p\in \{p_\text{sample},p_\text{neigh}\}}\text{Cost}(p)$
\STATE remove($D,p_\text{high}$)
\ELSE
\STATE $k \leftarrow k + 1$
\ENDIF
\ENDWHILE
\RETURN D
\end{algorithmic}
\end{algorithm}

For each point in input space, we are interested in retaining the lower bound of the cost of the generated datapoints. First note that we prefer to remove points in high-density regions, as in low-density regions there is little to throw away, and we may only hope that our data are accurate. We implicitly remove from high density regions by first uniformly sampling a point from our dataset. We then search for its nearest neighbour in the dataset based on a Euclidean distance. If this neighbour is within a distance $d$ from our sampled point, we remove the node of the two with the highest cost. Otherwise, we retain both points. This process is repeated until no points are removed for $k_m$ consecutive steps, after which we return the cleaned dataset. The procedure is outlined in Algorithm \ref{alg:clean_data}.

The main parameter in this algorithm, $d$, can be interpreted as a neighbourhood size. In low density regions, there will be no nearby points within distance $d$, and we will therefore never remove a point. In high-density regions, the algorithm will remove the highest cost datapoint of the two, frequently removing the biased one while retaining a good one. We can see the performance of this algorithm for different $d$ in Figure \ref{fig:distanceEffect}, which shows there is an optimal setting of $d$ that depends on the dataset. $d$ is a hyperparameter of our algorithm and has to be scaled empirically. In this case, this means fixing $d$, running the cleaning, fitting the function predictors, and then sampling new datapoints and assessing the error in cost and co-state parameters on the predictions. This is not ideal, but the proper evaluation metric (RRT performance) would come at additional computational cost. For our experimental problem, we found that a simple grid-search provides easy scaling of $d$.

\section{Machine learning considerations}
\label{sec:errors}

The proposed algorithm adds machine learning based approximations to  standard RRT, which intuitively might interfere with convergence of the algorithm. Therefore, we establish probabilistic completeness of our algorithm, by modifying the proof for the original RRT~\cite{LaValle2001}. We have not tried to make the proof outlined here as general as possible. The framework from \cite{Li2016} would potentially aid in that effort.

First, we assume there exists a solution to our planning problem, and that it is build out of a finite number of shorter trajectories. That is, the solution has $n$ waypoints $\mathcal{X}= \{x_0,x_1,\dots,x_n\}$. The controls between those waypoints are given by $\mathcal{U}=\{u_0,\dots,u_{n-1}\}$.

If $x_i$ is the most advanced waypoint currently in the tree, the chance of getting to the next node can be factored as
\begin{equation}
  P(\text{reach } x_{i+1}) = P(u_i|\text{expand }x_i)P(\text{expand } x_i)\label{eq:factors}
\end{equation}

Now if we can guarantee both factors are positive, i.e., $P(\text{expand } x_i)>0$ and $P(u_i|\text{expand } x_i)>0$, we get: $P(\text{reach } x_{i+1}) > 0$. This means the chance of getting to the next node of the solution is finite, so at some point the algorithm will get to the next node, and the next one, and so on. Therefore the algorithm will converge.

\subsection{Bounding the chance of picking the right action}

To ensure the chance of picking the right action is bounded from below, the number of possible inputs should be finite. Therefore we discretize the continuous input representation (the initial costate and the time duration of the trajectory). In the experiments, this is done by rounding them to 2-decimals.

Furthermore, a deterministic function approximator might not select the right steering input. Therefore we should use a probabilistic steering input, which could assign a higher probability to steering inputs closer to those suggested by the function approximator, but which gives at least some probability to each input.

In our experiments, the control parameters were sampled from truncated normals, with the bounds for each parameter specified by its sampled domain. The means are the value predicted by the learned model. The standard deviation $\sigma$ of the (non-truncated)-normal is a parameter of the algorithm.

When looking at practical performance, inaccurate prediction makes selecting the right action particularly difficult whenever the problem requires the RRT to very precisely reach a small region in state-space, for example when near the goal region. In experiments we found that we could reduce the time to get from `close to' the goal region to inside the goal region by increasing the standard deviation  when the predicion involves the goal state.

\subsection{Bounding the chance of picking the right node}

The chance of picking the right node is the volume of the statespace for which that node is the nearest node (as measured by the distance function used) divided by the total volume of the free state-space:
\begin{equation}
  P(x_i) = \frac{\text{\small Vol}(\{x\in\mathcal{X}_\text{free} |d(x_i,x)<d(x_e,x), \forall x_e\in \mathcal{E}\backslash x_i\})}{\text{Vol}(\mathcal{X}_\text{free})} \nonumber
\end{equation}
Since the volume of $\mathcal{X}_{free}$ is fixed and finite, we only need to make sure the numerator is non-zero. This should be done while taking into account that the cost-to-go function is piecewise continuous, with a discontinuity at $0$.

The first step is to impose that the distance function must always be larger than some positive constant times the Euclidean distance:
\begin{equation}
d(x,y) \geq c_l||x-y||_2\quad c_\text{lb}>0 \forall x,y \label{eq:conditionLB}
\end{equation}
There should also be something affecting an upper bound to the distance function. To construct this upper bound, we use the set  $\mathcal{G}(x)$, the largest connected set containing $x$ with points for which the distance to $x$ is bounded by $c_\text{ub}$ times the Euclidean distance:
\begin{align}
  \mathcal{G}(x)=\text{argmax} &\text{Vol}(\mathcal{S})\\
   &\text{s.t. }\mathcal{S}\subseteq \{y|d(x,y)\leq c_\text{ub}||x-y||_2\},\nonumber\\
  &\quad\;\; x \in \mathcal{S},\nonumber \\
  &\quad\;\;\mathcal{S}\text{ is connected}\nonumber
\end{align}
The upper bound condition then is as follows:
\begin{equation}
  \text{Vol}(\mathcal{G}(x))>c_\text{v}\quad\forall x \label{eq:conditionUB}
\end{equation}
Note that these conditions are not met in two frequently studied cases: \begin{enumerate*}\item  when the cost function is the integral of the squared input, \item when the system is not small time locally accessible, as happens for instance in underactuated systems.\end{enumerate*}

Take the largest ball $\mathcal{B}_\rho(x_i)$ centered around point $x_i$, such that $c_\text{ub}||x_i-y||\leq c_\text{lb}||x-y||$ for all $y$ in the ball, and all nodes $x$ in the tree. Based on simple Euclidean geometry, $\rho> 0$. Furthermore, by construction, the intersection $\mathcal{B}_\rho(x_i)\cap\mathcal{G}(x_i)$ has positive volume, and all points in that intersection are closer (by measure $d$) to node $x_i$ than to any other node in the tree. Together this shows that $P(\text{expand } x_{i})>0$.

If we had access to the true distance function, or an approximation of it that meets the conditions specified above, this would conclude the proof of convergence. However, learning algorithms are intended to generalize using interpolation, so may make large errors when extrapolating. This is especially true for Learning RRTs, for which this problem has not been identified in literature yet. For most machine learning, test data usually originate from the same data distribution (e.g. a picture, video, audio fragment or person characteristics) as the training data. However, in RRTs we \emph{uniformly} sample state-space, while we have confined our dataset to only contain short motion segments. Therefore, if we sample a new combination $(x_0,x_1)$, we have a reasonable chance of sampling outside of our dataset, where function approximation may make large errors, which might cause conditions \ref{eq:conditionLB} and \ref{eq:conditionUB} to be violated. Particularly, the approximated distance metric might greatly underestimate the cost-to-go from a certain node, causing that node to be incorrectly chosen for expansion.

In our implementation, we enforce the conditions by using a binary classifier that decides whether a query would yield a valid prediction of the cost and input parameters. We learn a function $\hat{V} : (\mathcal{X},\mathcal{X})\rightarrow \text{v}$, with $v\in [\texttt{true},\,\texttt{false}]$ which identifies when a combination of initial state and final state is valid (\texttt{true}), i.e. when the dataset $D$ covers that point in input-space. We implement a basic, but functional, $\hat{V}$-function that computes the summed distance to the nearest neighbours of the queried point to the points in the dataset, and rejects the query point if this sum becomes too large. An alternative approach relies on the notion that the dataset contains only short segments, meaning the final states should be reachable within a short period of time. The use of reachable sets to classify the validity of a distance metric in state-space RRT was already explored in ~\cite{Shkolink2009}.

Also, to avoid violations of conditions \ref{eq:conditionLB} and \ref{eq:conditionUB} by small approximation errors in the learned function, the predicted cost-to go is saturated at lower and upper bounds of $10^{\mp5}$. These bounds were chosen such that they enforce the conditions on the cost function, while their effect on the computation is negligable.

\section{Experiments}
\label{sec:Experiments}
To test our approach, we perform experiments on a relatively simple problem: pendulum swingup. This is a task on a single degree of freedom system in which a pendulum has to move from its stable equilibrium $(\theta,\omega) = (-\pi,0)$ to its unstable equilibrium $(0,0)$. The equations of motion for the pendulum are given in Section \ref{sec:dataGeneration}.

Data were generated and cleaned 10 times, to create 10 epochs, with 300 runs of the RRT algorithm per epoch.  The data for each epoch consist of 40000 simulations, which ended when the costs or norm of the state difference with the initial state exceeds \num{2} or \num{1.5}. Integration was done by the 4th order Runge-Kutta algorithm with a time step of \SI{0.01}{\second}. The initial position was uniformly sampled from $(-3\pi/2,\pi/2)$\si{\radian}, the initial velocity from $(-\pi,\pi)$\si{\radian\per\second}, and the initial costate sampled as described below. The data cleaning resolution $d$ equals 0.05. The data cleaning stopping parameter $k_\text{max}$ is set to 5000. The nearest neighbour fitting algorithm during the RRT takes $m=3$ nearest neighbours. Finally, the standard deviation of the sampling distribution $\sigma=\pi/4$ normally, and $\pi/2$ when the query involves the goal state.

To avoid projecting on the costate constraint (Eq. \ref{eq:costateConstraint}), which is computationally expensive for larger systems, we solve the constraint explicitly, i.e., uniformly sample the parameter $\phi\in(-\pi/2,3\pi/2)$ which sets the initial costate as follows:
\begin{align}
&\lambda_\theta = \tan(\phi)\nonumber\\
&\lambda_\omega = -\sin(\theta)+\text{sign}(\cos(\phi))\sqrt{\sin(\theta)^2 + 2 + \tan(\phi)\omega}\nonumber
\end{align}
If $\lambda_\omega$ has an imaginary part, the simulation is disregarded. The choice for the free parameter influences the sampling-density of the initial costates. Because the $\tan$-function has a low value on most of its domain, the initial costates tend to be small as well. This causes low initial torques, which is desired for the pendulum swing up.
The above parametrization can be generalized for input affine systems with a cost function that is quadratic in the input, a class that includes many mechanical systems.

\subsection*{Results}

\begin{figure}
\centering
\includegraphics[width=0.8\columnwidth]{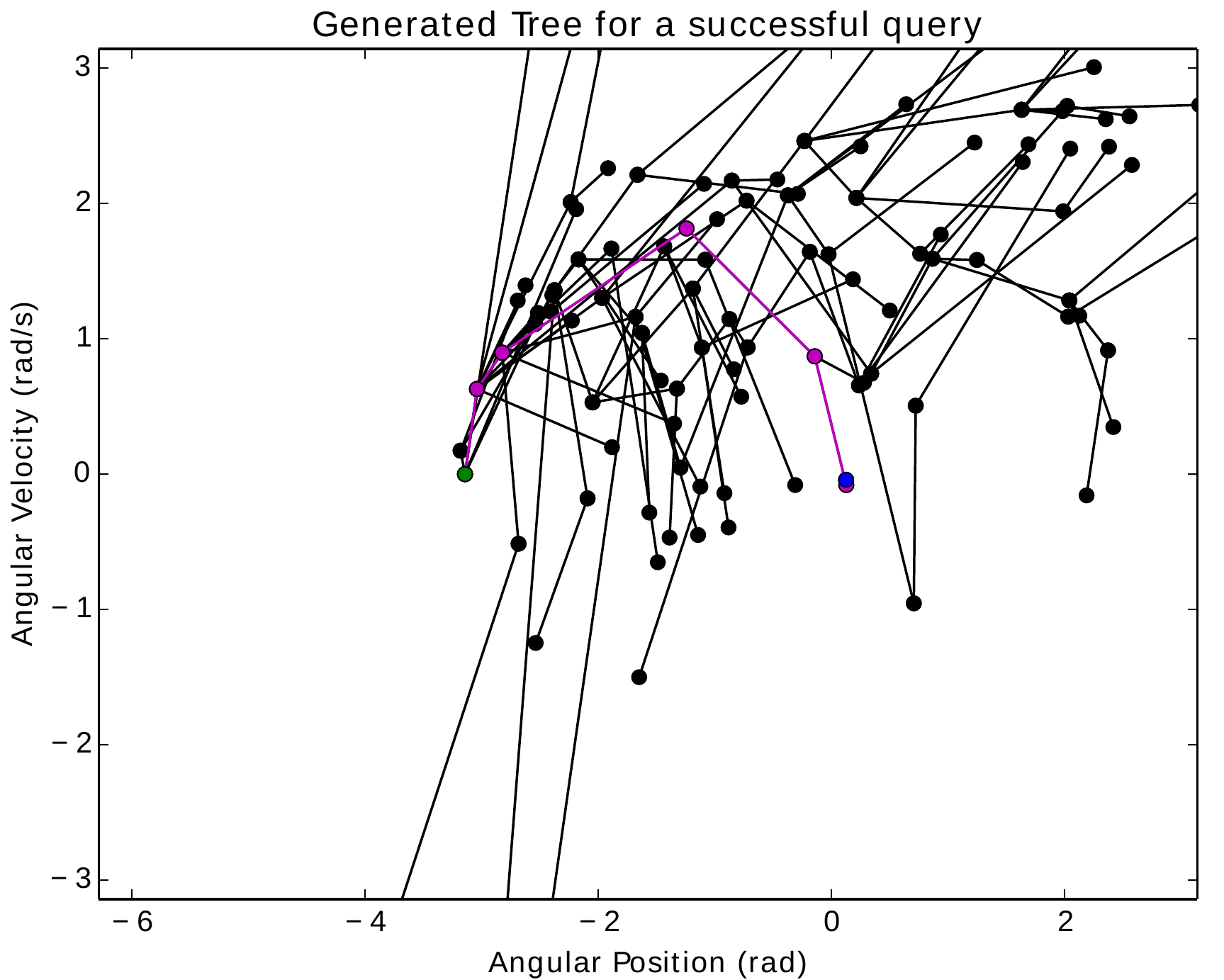}
\caption{The state-space coordinates of the tree-nodes of a succesful run of the algorithm, including their edges. Note that the trajectories between the nodes
are not straight lines. The trajectory from the initial point to the goal is highlighted.}
\label{fig:pendulum_swingup}
\end{figure}

Figure \ref{fig:pendulum_swingup} shows a typical run of the algorithm. Three important points can be seen from this figure. First, the algorithm neatly expands through state-space. While it favours expanding along the circular paths that correspond to low input trajectories, it expands nodes in all feasible directions. This is an indication that the distance metric works properly. Particularly, this contrasts with trees that are grown without appropriate distance metric, which tend to have many nodes clustered. Secondly, the figure highlights the approximation errors that still exists: the edges that lead to invisible nodes all had target nodes that would have been inside the graph. These end-nodes are now outside of the graph, which implies that the target node was not reached exactly. Finally, we see that the algorithm has tried to reach the goal-state a number of times from the same node. Here the effect of the added randomness becomes clear, because the attempts are spread out sufficiently, such that the goal is eventually reached.

\begin{figure}
\centering
\includegraphics[width=\columnwidth]{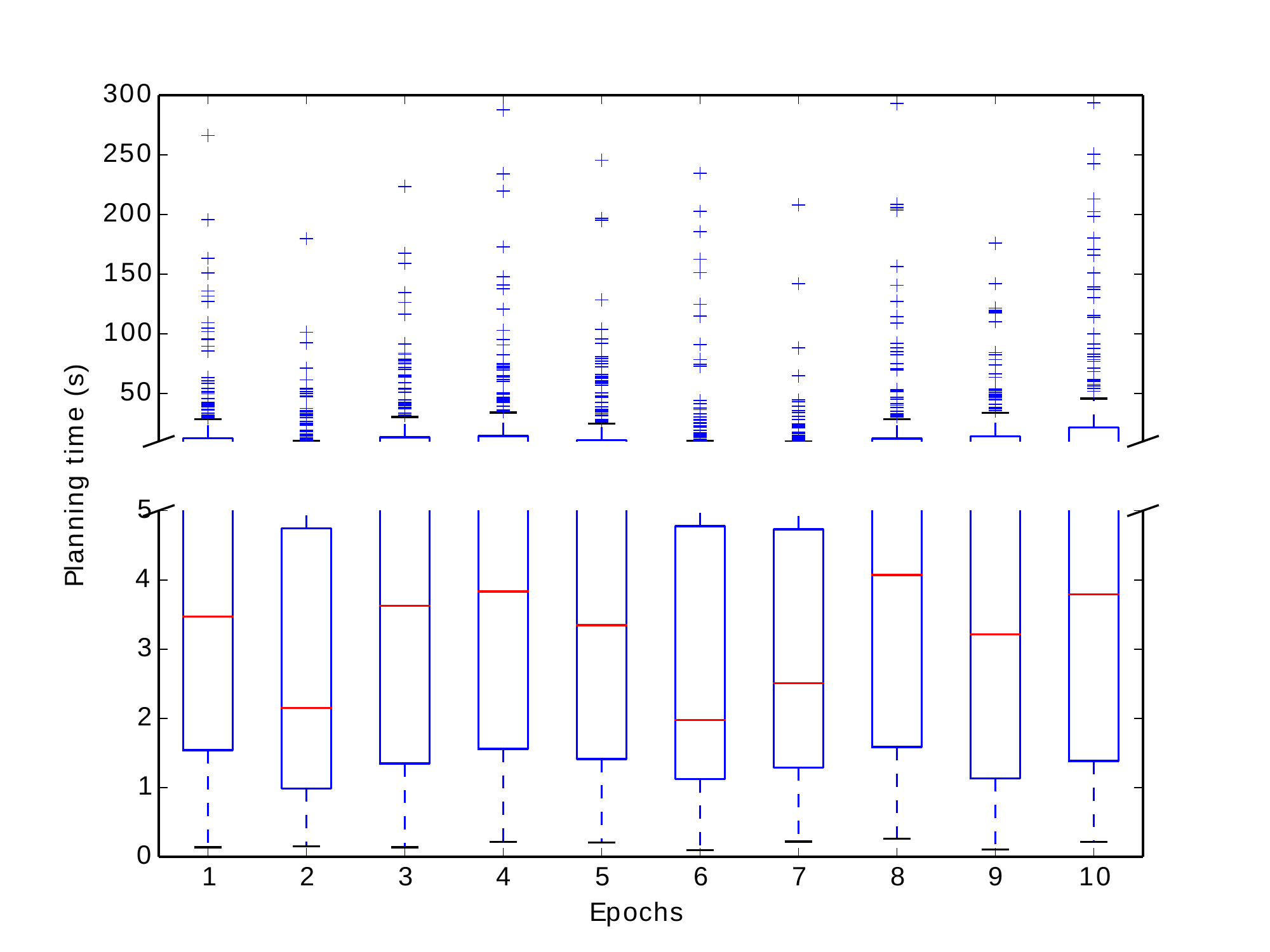}
\caption{Boxplot of the planning times, separated by epoch. It is visible that the planning time has the same variation in each epoch, indicating that the generation and cleaning of the data is performed robustly.}
\label{fig:times}
\end{figure}

The experiments were performed on a MacBook with Intel(R) Core(TM) i5-3210M CPU 2.50GHz processor and 8GB of RAM, running Ubuntu Linux 14.04. All the relevant code is written in Python. Figure \ref{fig:times} shows the variation in computation times for each epoch separately. The computation time does not change much between epochs, indicating that the data generation and cleaning are robust against random perturbations. Furthermore, it suggests that using multiple datapoints from a single simulation does not deteriorate the quality (i.i.d.-ness) of the dataset. The median time to reach the target over all samples was 2.36 seconds, more than 10 times faster than \citep{Bharatheesha2014} on the same hardware.

The simulations and data cleaning in this algorithm took a total of approximately \SI{25}{\minute} per epoch. This also is an order of magnitude faster than the algorithm from \citep{Bharatheesha2014}\footnote{The cited paper does not report the offline computation time. However, the authors of that paper overlap with the authors of this paper, so we know that the offline computation took nearly a week.}.

The performance of the optimal control function approximation is assessed using the mean squared error between the target state and the final state attained by using the predicted costate. The median of this error over all the epochs is 0.11. Furthermore, the approximation quality is indirectly measured by the number of nodes needed to reach the target. The median over all runs is \num{84} nodes, with a standard deviation of \num{180} nodes, which is about 30\% smaller (better) than the previous algorithm~\citep{Bharatheesha2014}. A slight decrease is expected, as the problem no longer requires the pendulum to swing back and forth to reach the final position. The decrease is therefore best interpreted as a roughly equal performance of the distance metric and optimal trajectory functions. This equal performance is obtained even though the optimal trajectory now uses function approximation.

\section{Discussion}
\label{sec:discussion}
The algorithm introduced in this paper allows learning of not only the distance metric, but also the steering input. As proof of concept, we tested our algorithm on a basic pendulum swing up problem, showing that it reduces the time spend both in the offline learning and in the online-computation by a factor of more than 10. This result is a large step towards making sampling based state-space planning in a practical setting feasible.

The main direction for future research is extending the algorithm for use on higher degree of freedom systems. The number of simulations required to learn the cost and costate functions is expected to grow rapidly with the number of degrees of freedom. Extension towards higher degrees of freedom would require a switch to different function approximators than the k-nearest neighbours used in this work.
To handle higher dimensions, it would be beneficial to have a higher sample efficiency of the dataset. The current data generation procedure samples uniformly from state and costate, which is likely inefficient. This might be improved by sampling new simulations based on the already obtained data, and the (partially) learned cost and costate functions. Similar ideas have been used in reinforcement learning \citep{sutton1998reinforcement}, \citep{levine2013guided}, and might be beneficial for use in Learning RRTs.
Finally, higher dimensional systems also require a more robust clean-up function. One improvement over the current cleaning function would be to not only compare trajectories based on their endpoints, but to explicitly take into account the distance in input-and-cost-to-go space. Alternatively, (deep) generative models \citep{goodfellow2016nips} allow to sample from complex, high-dimensional probability distributions. Using the approach from \citep{moerland2017learning}, we could retain all solutions while avoiding the averaging over solutions that is done in standard discriminative models.

A secondary direction for future research is to incorporate input bounds. Such bounds are readily incorporated in the indirect optimal control scheme, see~\citep{naidu2002}. However, there is an issue with the resulting costates: there can be an exact overlap between trajectories of a system with input bounds starting from different costates, at least for a finite time. Such overlapping trajectories cannot be handled by the basic learning and cleaning algorithms we used. Extending these algorithms, such that they can cope with such overlapping trajectories is an important theoretical and practical issue.

\section{Conclusion}
\begin{table}
\centering
\scriptsize{
\begin{tabular}{rlll}
\toprule
 & \bf Optimal Control & \bf Machine Learning & \bf RRT  \\
\midrule
\textbf{+} & Distance computation & Generalization & High dimensional \\
                  & Optimal local planner & Fast online prediction & Obstacle avoidance \\
\midrule
\textbf{-} & Costly computation\textsuperscript{\ref{sec:dataGeneration}}& Needs large dataset\textsuperscript{\ref{sec:dataGeneration}} & Needs distance metric \\
                    & Local optima$\rightarrow$bias\textsuperscript{\ref{sec:datasetCleaning}} & Needs unbiased data\textsuperscript{\ref{sec:datasetCleaning}} & Needs local planner\textsuperscript{\ref{sec:dataGeneration}}  \\
										&                           & Bad extrapolation\textsuperscript{\ref{sec:errors}}&                          \\
\bottomrule
\end{tabular}}
\caption{Benefits and challenges for RRT, Machine Learning and Optimal Control, which are combined in this paper. The challenges are marked with superscripts that refer to the sections in which they are treated.}
\label{tab:overview_table}
\end{table}

\label{sec:conclusion}
In this paper we described a general Learning RRT algorithm, and identified several problems with state-of the art versions. Table \ref{tab:overview_table} summarizes the parts that make up the algorithm, and their benefits and challenges.

We proposed the RRT-CoLearn Algorithm which addresses three problems of Learning RRT: \begin{enumerate*}
\item{By using indirect optimal control the number of parameters that describe the input is very small. The parameters of this function can thus be learned, alleviating the need for local planning in the online phase.}
\item{By using indirect optimal control, the data generation can be done much faster, as a numerical optimization is replaced by sampling.}
\item{An algorithm was proposed that removes the dataset bias caused by local optima.}
\end{enumerate*}

The RRT coLearn algorithm was tested on a pendulum swing up. It achieves a median planning time of \SI{2.4}{\second}, which is \num{10} times faster than the state-of the art learning algorithm for kinodynamic RRT.

\bibliographystyle{plainnat}
\bibliography{rrt_colearn_bibliography}
\end{document}